\documentclass[draftcls, 11pt, onecolumn]{IEEEtran}
\usepackage[left=1.25in,right=1.25in,top=1in,bottom=1in]{geometry}

\usepackage{tabularx,amsmath,amssymb,graphicx,paralist,subfigure,pdfpages,cite,algorithm,algpseudocode,paralist,multirow,comment}

\usepackage[autostyle]{csquotes}

\newtheorem{rmk}{Remark}

\def\ln{{\rm ln}}

\def\mc{\mathcal}
\def\mb{\mathbf}
\def\mbb{\mathbb}
\def\ra{\rightarrow}

\IEEEoverridecommandlockouts
\def\mbb{\mathbb}
\def\mb{\mathbf}
\def\mc{\mathcal}
\def\wh{\widehat}
\def\wt{\widetilde}
\def\ol{\overline}
\def\ul{\underline}

\def\bds{\boldsymbol}
\def\bth{\boldsymbol\theta}
\newcommand{\mn}[1]{{\left\vert\kern-0.25ex\left\vert\kern-0.25ex\left\vert\kern0.3ex #1 
		\kern0.3ex\right\vert\kern-0.25ex\right\vert\kern-0.25ex\right\vert}}

\begin{document}
	\title{\huge An introduction to decentralized stochastic optimization with gradient tracking}
	\author{Ran Xin$^\dagger$,  Soummya Kar$^\dagger$, and Usman A. Khan$^\ddagger$
		\\$^\dagger$Carnegie Mellon University, Pittsburgh, PA \hspace{1cm} $^\ddagger$Tufts University, Medford, MA
		\thanks{RX and SK are with the Electrical and Computer Engineering (ECE) department at Carnegie Mellon University, \texttt{\{ranx,soummyak\}@andrew.cmu.edu}. UAK is with the  department at Tufts University, \texttt{khan@ece.tufts.edu}. The work of SK and RX has been partially supported by NSF under award \#1513936. The work of UAK has been partially supported by NSF under awards \#1350264, \#1903972, and \#1935555. }
	}
	\maketitle

\vspace{-1.35cm}
\begin{abstract}
\vspace{-0.4cm}
Decentralized solutions to finite-sum minimization are of significant importance in many signal processing, control, and machine learning applications. In such settings, the data is distributed over a network of arbitrarily-connected nodes and raw data sharing is prohibitive often due to communication or privacy constraints. In this article, we review decentralized stochastic first-order optimization methods and illustrate some recent improvements based on gradient tracking and variance reduction, focusing particularly on smooth and strongly-convex objective functions. We provide intuitive illustrations of the main technical ideas as well as applications of the algorithms in the context of decentralized training of machine learning models.  
\end{abstract}
	
\section{Introduction}
In multi-agent networks and large-scale machine learning, when data is collected from geographically dispersed, resource-constrained devices, or when data is stored on different machines with limited communication capabilities, it is often desirable to seek scalable learning and inference solutions that do not require bringing, storing, and processing data at one single location. Besides, to leverage modern computational resources, such as computing clusters, advanced computational frameworks that are communication-efficient and able to explore data parallelism are particularly favorable. In this magazine article, we describe decentralized, consensus-based, stochastic first-order methods, which are particularly well-fitted to the aforementioned \textit{ad-hoc} and \textit{resource-constrained} scenarios. Specifically, in the context of first order methods, we show how a recently introduced gradient tracking technique provides a systematic approach to designing decentralized versions of centralized stochastic gradient methods. We then use recent advancements on variance reduction to derive decentralized methods that are particularly advantageous in problems where high-precision solutions are desired. To keep the exposition simple, we focus on smooth and strongly-convex objective functions, however, the algorithms are applicable to general convex and non-convex problems. To provide context, we start by briefly reviewing the corresponding optimization problems and their associated centralized solutions that commonly arise in the signal processing and machine learning literature.
	
\subsection{Empirical Risk Minimization}\label{sERM}
In parametric learning and inference problems, the goal of a typical machine learning system is to find a model~$g$, parameterized by a real vector~$\bds{\theta}\in\mathbb{R}^p$, that maps an input data point~$\mb{x}\in\mathbb{R}^{d_{\mb{x}}}$ to its corresponding output~$\mb{y}\in\mathbb{R}^{d_{\mb{y}}}$. The setup requires defining a loss function~$l(g(\bth;\mb{x,y}))$, which represents the loss incurred by the model~$g$ with para{}meter~$\bds{\theta}$ on the data~$(\mb{x},\mb{y})$. In the formulation of statistical machine learning, we assume that each data point~$(\mb{x},\mb{y})$ belongs to a joint probability distribution~$\mc{P}(\mb{x},\mb{y})$. Ideally, we would like to find the optimal model parameter~$\widetilde{\bds{\theta}}^*$ by minimizing the following \textit{risk (expected loss) function}~$\widetilde{F}(\bds{\theta})$:
\begin{equation*}
\mbox{P0:} \qquad\widetilde{\bds{\theta}}^* = \operatorname*{argmin}_{\bds{\theta}\in\mathbb{R}^p}
\widetilde{F}(\bds{\theta}),\qquad \widetilde{F}(\bds{\theta}) \triangleq 
\mathbb{E}_{(\mb{x},\mb{y})\sim\mc{P}(\mb{x},\mb{y})}
l(g(\bth;\mb{x,y})).
\end{equation*} 
However, the true distribution~$\mc{P}(\mb{x},\mb{y})$ is often hidden or intractable in practice. In supervised machine learning, one usually has access to a large set of training data points~$\{\mb{x}_i,\mb{y}_i\}_{i=1}^m$, which can be considered as the independent and identically distributed (i.i.d.) realizations from the distribution~$\mc{P}(\mb{x},\mb{y})$. The average of the losses incurred by the model~$\bds{\theta}$ on a finite set of the training data~$\{\mb{x}_i,\mb{y}_i\}_{i=1}^N$, known as the \textit{empirical risk}, thus serves as an appropriate surrogate for the risk function~$\widetilde{F}(\bds{\theta})$. Formally, one may find the optimal~model parameter~$\bds{\theta}^*$ by solving the following \textit{empirical risk minimization}, instead of P0: 
\begin{equation}
\mbox{P1:}\qquad\bds{\theta}^* = \operatorname*{argmin}_{\bds{\theta}\in\mathbb{R}^p}F(\bds{\theta}),\qquad F(\bds{\theta})\triangleq\frac{1}{N}\sum_{i=1}^{N}l(g(\bth;\mb{x}_i,\mb{y}_i)) \triangleq \frac{1}{N}\sum_{i=1}^{N} f_{i}(\bds{\theta}).
\end{equation}
This finite-sum formulation captures a wide range of supervised learning problems, e.g., least-square regression, logistic regression, support vector machines, and deep neural networks~\cite{PRML}.

This article focuses on smooth and strongly-convex objectives, defined as follows. An \textit{$L$-smooth} function~$f: \mathbb{R}^p\ra\mathbb{R}$ is such that~$\forall\bds{\theta}_1, \bds{\theta}_2\in\mbb{R}^p$ and for some positive constant~$L>0$, we have $$\|\mb{\nabla} f(\bds{\theta}_1)-\mb{\nabla} f(\bds{\theta}_2)\|_2\leq L\|\bds{\theta}_1-\bds{\theta}_2\|_2.$$ A \textit{$\mu$-strongly-convex} function~$f: \mathbb{R}^p\ra\mathbb{R}$ is such that~$\forall\bds{\theta}_1, \bds{\theta}_2\in\mbb{R}^p$ and for some positive constant~$\mu>0$, we have $$f(\bds{\theta}_2)\geq f(\bds{\theta}_1)+\nabla f(\bds{\theta}_1)^\top(\bds{\theta}_2-\bds{\theta}_1)+\frac{\mu}{2}\|\bds{\theta}_1-\bds{\theta}_2\|_2^2.$$ We define~$\mc{S}_{\mu,L}$ as the class of functions that are~$\mu$-strongly-convex and~$L$-smooth. We note that if each~$f_i\in\mc{S}_{\mu,L}$, then~$F\in\mc{S}_{\mu,L}$, and~$F$ has a unique global minimum denoted as~$\bth^*$. For any~$F\in\mc{S}_{\mu,L}$, we note that~$L\geq\mu$, and we define~$\kappa\triangleq\frac{L}{\mu}$ as the condition number of~$F$~\cite{nesterov_book}; clearly,~$\kappa\geq1$; a function with a large condition number is said to be ill-conditioned.

\subsection{Stochastic Gradient Descent}\label{sec_sgd}
Stochastic Gradient Descent (SGD) is a simple yet powerful method that has been extensively used to solve the empirical risk minimization problem P1. SGD, in its simplest form, starts with an arbitrary~$\bds{\theta}_0\in\mathbb{R}^p$ and performs the following iterations to asymptotically learn~$\bth^*$ as~$k\ra\infty$:
\begin{align}\bds{\theta}_{k+1} = \bds{\theta}_k - \alpha_k \cdot  \nabla f_{s_k}(\bds{\theta}_k), \qquad k\geq0, \label{csgd}
\end{align}
where~$s_k$ is chosen randomly from~$\{1,\cdots,N\}$ and~$\{\alpha_k\}_{k\geq 0}$ is a sequence of positive step-sizes. Comparing to batch gradient descent where the descent direction~$\nabla F(\bth_k)$ is computed from the entire batch of data, SGD's descend direction is the gradient of a randomly sampled component function. SGD is thus computationally efficient as it evaluates one gradient (easily extendable to more than one randomly selected functions) at each iteration and is a popular alternative in problems with a large number of high-dimensional training data samples and model parameters.

We note that the stochastic gradient~$\nabla f_{s_k}(\bth_k)$ is an unbiased estimate of the batch gradient~$\nabla F(\bth_k)$, i.e.,~$\mathbb{E}_{s_k}[\nabla f_{s_k}(\bds{\theta}_k)|\bth_k] = \nabla F(\bth_k)$. Under the assumptions that each~$f_i\in\mc{S}_{\mu,L}$ and each stochastic gradient~$\nabla f_{s_k}(\bds\theta_k)$ has bounded variance, i.e.,~$$\mathbb{E}_{s_k}\left[\left\|\nabla f_{s_k}(\bds\theta_k)-\nabla F(\bds\theta_k)\right\|_2^2 |\bds\theta_k\right] \leq{\sigma}^2,\qquad~\forall k,$$ it can be shown that with a constant step-size~$\alpha\in\left(0,\frac{1}{L}\right]$,~$\mathbb{E}\left[\|\bds{\theta}_k-\bds{\theta}^*\|_2^2\right]$ decays geometrically, at the rate of~$\left(1-\mu\alpha\right)^k$, to a neighborhood of~$\bth^*$. Formally, we have~\cite{OPT_ML},
\begin{align}\label{sgd_conv}
\mathbb{E}\left[\|\bds{\theta}_k-\bds{\theta}^*\|_2^2\right]
\leq (1-\mu\alpha)^k + \frac{\alpha{\sigma}^2}{\mu},\qquad \forall k\geq0.
\end{align}
This steady-state error or the inexact convergence is due to the fact that~$\nabla f_{s_k}(\bds{\theta}^*) \neq 0$, in general, and the step-size is a constant. A diminishing step-size,~$\mc{O}(\frac{1}{k})$, overcomes this issue and leads to exact convergence albeit at slower rate. For example, with~$\alpha_k = \frac{1}{\mu (k+1)}$, we have~\cite{OPT_ML},
\begin{align}\label{sgd_conv_2}
\mathbb{E}\left[\|\bds{\theta}_k-\bds{\theta}^*\|_2^2\right]
\leq \frac{\max\left\{\frac{2{\sigma}^2}{\mu^2},\|\bth_0-\bth^*\|_2^2\right\}}{k+1},\qquad\forall k\geq0.
\end{align}
In other words, to reach an~$\epsilon$-accuracy of the optimal solution~$\bth^*$,~i.e.,~$\mathbb{E}\left[\|\bds{\theta}_k-\bds{\theta}^*\|^2\right]\leq\epsilon$, SGD (with decaying step-sizes) requires~$\mc{O}\left(\frac{1}{\epsilon}\right)$ component gradient evaluations. 

\subsection{Variance-Reduced Stochastic Gradient Descent}\label{VRSGD}
In practice, a successful implementation of SGD relies heavily on the tuning of the step-sizes and a decaying step-size sequence~$\{\alpha_k\}_{k\geq0}$ has to be carefully chosen due to the potential large variance in SGD, i.e., the sampled gradient~$\nabla f_{s_k}(\bds{\theta}_k)$ at~$\bds{\theta}_k$ can be very far from the batch gradient~$\nabla F(\bds{\theta}_k)$. In recent years, certain Variance-Reduction (VR) techniques have been developed towards addressing this issue~\cite{SAG,SAGA,SVRG,SARAH}. The key idea here is to design an iterative estimator of the batch gradient whose variance progressively decays to zero as~$\bth_k$ approaches~$\bth^*$. Benefiting from this, VR methods have a low per-iteration computation cost, a key feature of SGD, and, at the same time, converge geometrically to the exact solution~$\bth^*$ as the batch gradient descent. Different constructions of the aforementioned gradient estimator lead to different VR methods~\cite{SAG,SAGA,SVRG,SARAH}. We focus on two popular VR procedures next.

\textbf{SAGA~\cite{SAGA}: }The SAGA method starts with an arbitrary~$\bth_0\in\mathbb{R}^p$ and maintains a table that stores all component gradients~$\{\nabla f_i(\wh{\bth}_i)\}_{i=1}^N$, where~$\wh{\bth}_{i}$ denotes the most~recent iterate at which~$\nabla f_i$ was evaluated, initialized with~$\{\nabla f_i(\bth_0)\}_{i=1}^N$. At every iteration~$k\geq0$, SAGA chooses~$s_k$ randomly from~$\{1,\ldots,N\}$ and performs the following two updates:
\begin{align}
    \mb{g}_k = \nabla f_{s_k}(\bth_k) - \nabla f_{s_k}(\wh{\bth}_{s_k}) + \frac{1}{N}\sum_{i=1}^N\nabla f_i(\wh{\bth}_{i}),\ \qquad \bth_{k+1}=\bth_k-\alpha\cdot \mb{g}_k.
\end{align}
Subsequently, the entry~$f_{s_k}(\wh{\bth}_{s_k})$ in the gradient table is replaced by~$\nabla f_{s_k}\big(\bth_{k}\big)$, while the other entries remain unchanged, see Remark~\ref{VRdisc} below. Under the assumption that each~$f_i\in\mc{S}_{\mu,L}$, it can be shown that with~$\alpha = \frac{1}{3L}$, we have~\cite{SAGA},
\begin{align}
\mathbb{E}\left[\left\|\bth_k-\bth^*\right\|_2^2\right]\leq C\left(1-\min\left\{\frac{1}{4N},\frac{\mu}{3L}\right\}\right)^k,\qquad \forall k\geq0,
\end{align}
where~$C>0$ is some constant. In other words, SAGA achieves~$\epsilon$-accuracy of the optimal solution~$\bth^*$ with~$\mc{O}\left(\max\{N,\kappa\}\log\frac{1}{\epsilon}\right)$ component gradient evaluations, where recall that~$\kappa=\frac{L}{\mu}$ is the condition number of the global objective~$F$. 
Indeed, SAGA has a non-trivial storage cost of~$\mc{O}\left(Np\right)$ due to the gradient table required, which can be reduced to~$\mc{O}(N)$ for certain problems of interest by exploiting the structure of the objective functions~\cite{SAGA}. 

\textbf{SVRG~\cite{SVRG}: }Instead of storing the gradient table, SVRG achieves variance reduction by computing the batch gradient periodically and can be interpreted as a ``double-loop" method described as follows. The outer loop of SVRG, indexed by~$k$, updates the estimate~$\{\bth_k\}_{k\geq0}$ of~$\bth^*$. At each outer iteration~$k$, SVRG computes the batch gradient~$\nabla F(\bds{\theta}_k)$ and executes a finite number~$T$ of inner stochastic gradient iterations, indexed by~$t$: set~$\ul\bth_{0} = \bth_k$ and for~$t = 0,\cdots,T-1$, 
\begin{align}\label{csvrg}
\mb{v}_{t} = \nabla f_{s_{t}}(\ul\bth_{t}) - \nabla f_{s_{t}}(\ul\bth_{0}) + \nabla F(\ul\bth_{0}), \qquad
\ul\bth_{t+1} = \ul\bth_{t} - \alpha \cdot \mb{v}_{t}, 
\end{align}
where~$s_{t}$ is randomly selected from~$\{1,\cdots,N\}$, see Remark~\ref{VRdisc}. After the inner loop completes,~$\bds{\theta}_{k+1}$ can be updated in a few different ways; applicable choices include setting~$\bds{\theta}_{k+1}$ as~$\ul\bth_{T}$,~$\tfrac{1}{T}\sum_{t=0}^{T-1}\ul\bth_{t}$, or a uniform random selection from the inner loop updates~$\{\ul\bth_{t}\}_{t=0}^{T-1}$. Under the assumption that each~$f_i\in\mc{S}_{\mu,L}$, it can be shown that with~$\bds{\theta}_{k+1}=\tfrac{1}{T}\sum_{t=0}^{T-1}\ul\bth_{t}$,~$\alpha = \frac{1}{10L}$, and~$T = 50\kappa$, we have~\cite{SVRG},
\begin{align}
\mathbb{E}\left[\left\|\bds{\theta}_{k}-\bth^*\right\|^2\right]\leq D \cdot 0.5^k,\qquad \forall k\geq 0,
\end{align}
where~$D$ is some positive constant. That is to say, SVRG achieves~$\epsilon$-accuracy with~$\mc{O}(\log\frac{1}{\epsilon})$ outer-loop iterations. We further note that each outer-loop update requires~$N+2T$ component gradient evaluations according to~\eqref{csvrg}. Therefore, SVRG achieves~$\epsilon$-accuracy with~$\mc{O}\left((N+\kappa)\log\frac{1}{\epsilon}\right)$ component gradient evaluations, which is comparable to the convergence rate of SAGA.

\begin{rmk}\label{VRdisc}
One can verify that both~$\mb{g}_k$ and~$\mb{v}_{t}$ are unbiased\footnote{In fact, other variance reduction schemes such as SAG~\cite{SAG} and SARAH~\cite{SARAH} are based on biased gradient estimators and exhibit similar performance as SAGA and SVRG for smooth and strongly-convex functions.} estimators of the corresponding batch gradient, i.e.,~$\mathbb{E}_{s_k}[\mb{g}_k|\bth_k] = \nabla F(\bth_k)$ and~{\color{black}$\mathbb{E}_{s_{t}}[\mb{v}_{t}|\ul{\bth}_{t},\ul\bth_{0}] = \nabla F(\ul{\bth}_{t})$.} Therefore,~$\mb{g}_k$ and~$\mb{v}_{t}$ can be viewed as generalized stochastic gradients, which approach the batch gradient as their variance diminishes. VR methods~\cite{SAG,SAGA,SVRG,SARAH} are popular solutions for large-scale empirical risk minimization, particularly when high-accuracy solutions are desired. When low-precision solutions suffice, SGD can be quite effective as its convergence rate is independent of the sample size~$N$ and it typically makes fast progress in its early stage. That SGD does not depend on~$N$ is a remarkable feature, but it comes at a price of a complexity that is proportional to~$\sigma^2$ according to~\eqref{sgd_conv_2}; the convergence of VR methods, on the contrary, is independent of~$\sigma^2$. 
\end{rmk}

In the rest of this article, we show how to adapt SGD and VR methods in the decentralized optimization framework. Section~\ref{secPF} describes the corresponding optimization problem over a network of nodes. In Section~\ref{S2}, we extend centralized SGD to the decentralized setting and show that an appropriate decentralization is achieved with the help of a certain gradient tracking technique. Subsequently, in Section~\ref{S3}, we describe recent advancement in decentralized methods that combine gradient tracking and variance reduction. Section~\ref{S5} provides detailed numerical illustrations while Section~\ref{S4} summarizes certain extensions and secondary aspects of the corresponding problems. Finally, Section~\ref{S6} concludes the paper.

\section{Problem Formulation: Decentralized Empirical Risk Minimization}\label{secPF}
\begin{figure*}[!h]
\centering
\subfigure{\includegraphics[width=2.2in]{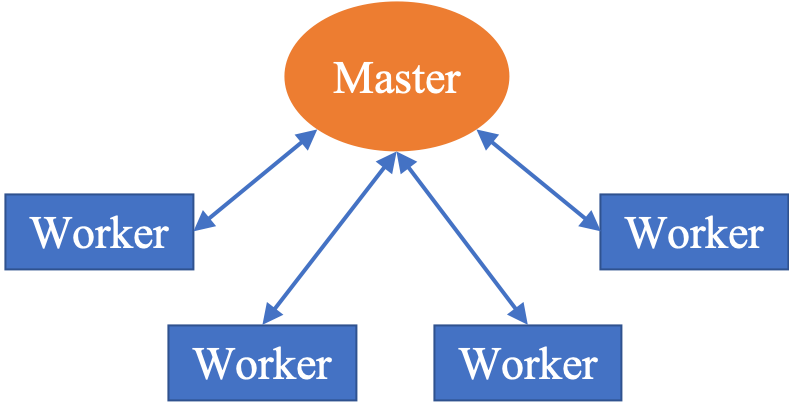}}\hspace{2cm}
\subfigure{\includegraphics[width=2.2in]{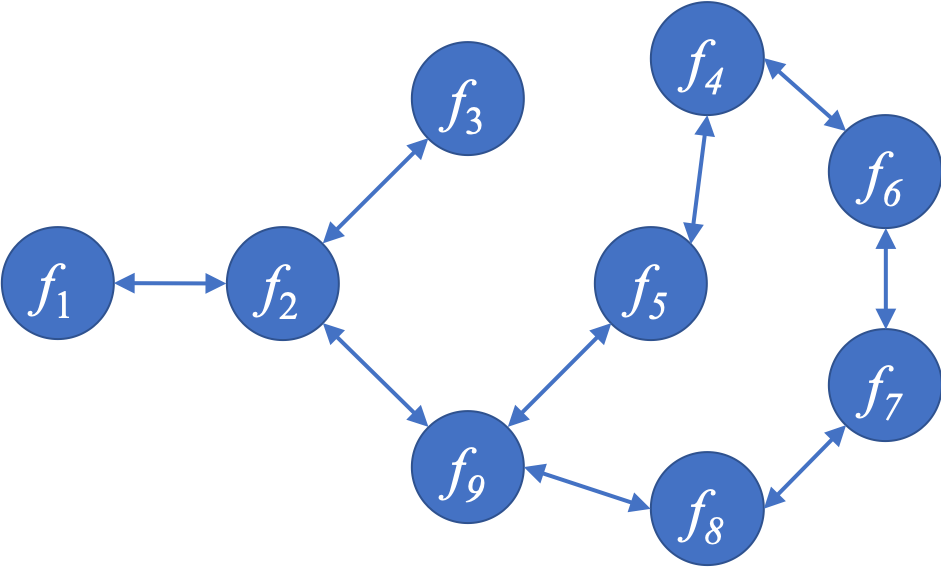}}
\caption{(Left) A master-worker network. (Right) Decentralized optimization in peer-to-peer networks.}
\label{master_worker}\label{decentralized}
\end{figure*}

In this magazine article, our focus is on the solutions for optimization problems that arise in peer-to-peer decentralized networks. In traditional master-worker architectures, see Fig.~\ref{decentralized} (left), a central node acts like a master that coordinates communications with all workers. In peer-to-peer networks, however, no such master node or central coordinator is available and each node is only able to communicate with its immediate neighbors, see Fig.~\ref{decentralized} (right), according to an arbitrary and ad hoc topology. The canonical form of decentralized optimization problems can be described as follows. Consider~$n$ nodes, such as machines, devices, or decision-makers, that communicate over an \textit{arbitrary undirected and connected graph}~$\mc{G}=(\mc{V},\mc{E})$, where~$\mc{V}=\{1,\cdots,n\}$ is the set of nodes, and~$\mc{E}\subseteq\mc{V}\times\mc{V}$ is the set of edges, i.e., a collection of ordered pairs~$(i,r),i,r\in\mc{V}$, such that nodes~$i$ and~$r$ can exchange information. Following the discussion in Section~\ref{sERM}, each node~$i$ holds a private and \textit{local risk function},~$\wt f_i:\mathbb{R}^p\ra\mathbb{R}$, not accessible by any other node in the network. 
The decentralized risk minimization problem can thus be defined as 
\begin{align}
\mbox{P2:}
\quad\widetilde\bth^* = \operatorname*{argmin}_{\bds{\theta}\in\mathbb{R}^p}\wt F(\bds{\theta}),\qquad \wt F(\bds\theta)\triangleq\frac{1}{n}\sum_{i=1}^n\wt f_i(\bds\theta).\nonumber
\end{align}

As in the centralized case with Problem P0, the underlying distributions may not be available or tractable at any node, we thus employ local empirical risks as appropriate surrogates. Specifically, we consider each node~$i$ to be a computing resource that stores/collects a local batch~$m_i$ of training data samples that are possibly private (not shared with other nodes) and the corresponding local empirical risk function is decomposed over the local data samples as~$f_i\triangleq\frac{1}{m_i}\sum_{j=1}^{m_i}f_{i,j}$. The goal of the networked nodes is to \textit{agree} on the \textit{optimal} solution of the
following \textit{decentralized empirical risk minimization} problem:
\begin{align}
\mbox{P3:}
\quad\bds{\theta}^* = \operatorname*{argmin}_{\bds{\theta}\in\mathbb{R}^p}F(\bds{\theta}),\qquad F(\bds\theta)=\frac{1}{n}\sum_{i=1}^nf_i(\bds\theta)\triangleq\frac{1}{n}\sum_{i=1}^n\:\left(\frac{1}{m_i}\sum_{j=1}^{m_i}f_{i,j}(\bds\theta)\right).\nonumber
\end{align}
The rest of this article is dedicated to the solutions of the above problem. 

\section{Decentralized Consensus-based Stochastic Optimization}\label{S2}
We now consider decentralized iterative algorithms to solve Problem P3.
At each node~$i$ given the current estimate~$\bds\theta_k^i$ of~$\bth^*$ at iteration~$k$, related decentralized consensus-based algorithms typically involve the following steps at each node~$i$:
\begin{enumerate}
\item Choose a local mini batch by sampling one or more component gradients from~$\{\nabla f_{i,j}\}_{j=1}^{m_i}$ evaluated at the current local iterate~$\bth_k^i$;
\item Fuse information with the available neighbors;
\item Compute~$\bds\theta_{k+1}^i$ according to a specific optimization protocol.
\end{enumerate}
Recall that each node in the network only communicates with a few nearby nodes and only has partial knowledge of the global objective, see Fig.~\ref{decentralized} (right). Due to this limitation, an information propagation mechanism is required that disseminates local information over the entire network. Consensus-based optimization, as its name suggests,  has two key components: (i) agreement or consensus: all nodes must agree on the same state, i.e.,~$\bds\theta_k^i\ra\bds\theta_{cons},\forall i$; and, (ii) optimality: the agreement should be on the minimizer of the global objective~$F$, i.e.,~$\bds\theta_{cons} = \bds\theta^*$. Average-consensus algorithms are information fusion protocols that enable each node to appropriately combine the vectors received from its neighbors and to agree on the average of the initial states of the nodes. They thus naturally serve as basic building blocks in decentralized optimization, added to which is a gradient correction that locally steers the agreement to the global minimizer. 

To describe average-consensus, we first associate the undirected and connected graph~$\mc{G}$ with a primitive, symmetric, and doubly-stochastic~$n\times n$ weight matrix~$W=\{w_{ir}\}$, such that~$w_{ir}\neq0$ for each~$(i,r)\in\mc{E}$. Clearly, we have~$W=W^\top$ and~$W\mb{1}_n=\mb{1}_n$, where~$\mb 1_n$ is the column vector of~$n$ ones. There are various ways of constructing such weights in a decentralized manner. Popular choices include the Laplacian and Metropolis weights, see~\cite{tutorial_nedich} for details. Average-consensus~\cite{consensus_Xiao} is given as follows. Each node~$i$ starts with some vector~$\bth_0^i\in\mathbb{R}^p$ and updates its state according to~$\bth_{k+1}^i = \sum_{r\in\mc{N}_i}w_{ir} \bth_{k}^r$,~$\forall k\geq0$. It can be written in a matrix form as
\begin{equation}\label{average_consensus_undirected_matrix}
\bds\theta_{k+1} = (W\otimes I_p) \bds\theta_{k},
\end{equation} 
where~$\bds\theta_k=[{\bth_k^1}^\top,\cdots,{\bth_k^n}^\top]^\top$. Since~$W$ is primitive and doubly-stochastic, from the Perron-Frobenius theorem, we have~$\lim\limits_{k\ra\infty} W^k = \frac{1}{n}\mb{1}_n\mb{1}_n^\top$ and~$
\lim\limits_{k\ra\infty}\bds\theta_{k} = (W\otimes I_p)^k\bds\theta_0 = (\mb{1}_n\otimes I_p) \ol{\bth}_0,$
where~$\ol{\bth}_0 \triangleq \frac{(\mb{1}_n^\top\otimes I_p)\bth_0}{n}$, at a geometric rate of~$\lambda^k$, and~$\lambda\in(0,1)$ is the second largest eigenvalue of~$W$. That is to say, the protocol in~\eqref{average_consensus_undirected_matrix} enables an agreement across all nodes on the average~$\ol{\bth}_0$ of their initial states, at a geometric rate. With the agreement protocol in place, we next introduce decentralized gradient descent and its stochastic variant that build on top of average-consensus.
	
\subsection{Decentralized Stochastic Gradient Descent (DSGD)}
Recall that our focus is to solve Problem P3 in a decentralized manner, when the nodes exchange information over an arbitrary undirected graph.
A well-known solution to this problem is Decentralized Gradient Descent (DGD)~\cite{DGD_nedich,diffusion_Chen}, described as follows. Each node~$i$ starts with an arbitrary~$\bth_i^0\in\mathbb{R}^p$ and performs the following update:
\begin{equation}\label{DGD}
\bds\theta_{k+1}^i = \sum_{r\in\mc{N}_i}w_{ir}\bds\theta_{k}^r - \alpha_k\nabla f_i\left(\bds\theta_k^i\right),\qquad k\geq 0.
\end{equation}
Indeed, DGD adds to average-consensus a local gradient correction based on the local batch data, i.e., all~$f_{i,j}$'s, and is the prototype of many consensus-based optimization protocols. To understand the iterations of DGD, we write them in a matrix form. Let~$\bds\theta_k$ and~$\nabla\mb{f}(\bds\theta_k)$ collect all local estimates and gradients, respectively, i.e.,~$\bds\theta_k=[{\bds\theta_k^1}^\top,\cdots,{\bds\theta_k^n}^\top]^\top$ and~$\nabla\mb{f}(\bds\theta_{k})\triangleq[{\nabla f_1(\bds\theta_k^1)}^\top,\cdots,{\nabla f_n(\bds\theta_k^n)}^\top]^\top$, both in~$\mathbb{R}^{np}$. Then DGD can be compactly written as
\begin{equation}\label{DGD_matrix}
\bds\theta_{k+1} = (W\otimes I_p)\bds\theta_k - \alpha_k\nabla\mb{f}(\bds\theta_k).
\end{equation}
We further define the average~$\ol{\bds\theta}_k \triangleq \frac{1}{n}(\mb{1}_n^\top\otimes I_p)\bds\theta_k$ of the local estimates at time~$k$ and multiply both sides of~\eqref{DGD_matrix} by~$(\mb{1}_n^\top\otimes I_p)$ to obtain:
\begin{equation}\label{DGD_average}
\ol{\bds\theta}_{k+1} = \ol{\bds\theta}_k - \alpha_k\frac{(\mb{1}_n^\top\otimes I_p)\nabla\mb{f}\left(\bds\theta_k\right)}{n}.
\end{equation} 
Based on~\eqref{DGD_matrix} and~\eqref{DGD_average}, we note that the consensus matrix~$W$ makes the estimates~$\{\bds\theta_k^i\}_{i=1}^n$ at the nodes approach to their average~$\ol{\bds\theta}_k$, while the averaged gradient~$\frac{(\mb{1}_n^\top\otimes I_p)\nabla\mb{f}\left(\bds\theta_k\right)}{n}$ steers the average~$\ol{\bds\theta}_k$ towards the minimizer~$\bth^*$ of~$F$. The overall protocol thus ensures agreement and optimality, the two key components of decentralized optimization as we described before. 

\begin{algorithm}[!h]
		\caption{DSGD: At each node~$i$}
		\begin{algorithmic}[1]
			\Require $\bds\theta_i^0$,~$\{\alpha_k\}_{k\geq0}$,~$\{w_{ir}\}_{r\in\mc N_i}$. 
			\For{$k= 0,1,2,\cdots$}
			\State\textbf{Choose}~$s_k^i$ uniformly at random in~$\{1,\cdots,m_i\}$ 
			\State \textbf{Compute} the local stochastic gradient~$\nabla f_{i,s_k^i}(\bth_k^i)$.
			\State \textbf{Update}:~$\bds\theta_{k+1}^i = \sum_{r\in\mc{N}_i}w_{ir}\bds\theta_k^r
			- \alpha_k\nabla f_{i,s_k^i}(\bth_k^i)$
			\EndFor
		\end{algorithmic}
	\end{algorithm}
DGD is a simple yet effective method for various decentralized learning problems. To make DGD efficient for large-scale decentralized ERM, where each~$m_i$ is very large, Refs.~\cite{DSGD_nedich,diffusion_Chen} derive a stochastic variant, known as \textit{Decentralized Stochastic Gradient Descent (DSGD)}, by substituting each local batch gradient with a randomly sampled component gradient. DSGD is formally described in Algorithm 1. Assuming that each~$f_{ij}\in\mc{S}_{\mu,l}$ and each local stochastic gradient has bounded variance, i.e.,~$\mathbb{E}_{s_k^i}\left[\left\|\nabla f_{i,s_k^i}(\bds\theta_k^i)-\nabla f_i(\bds\theta_k^i)\right\|_2^2 |\bds\theta_k^i\right] \leq\sigma^2,\forall i,k$, we have~\cite{DSGD_Yuan}: under a constant step-size,~$\alpha_k=\alpha\in\left(0,\mc{O}\left(\tfrac{(1-\lambda_w)\mu}{L^2}\right)\right],\forall k$, $\mathbb{E}[\|\bds\theta_k^i-\bds\theta^*\|_2^2]$ decays at a geometric rate of~$\left(1-\mc{O}(\mu\alpha)\right)^k$ to a neighborhood of~$\bth^*$ such that 
\begin{equation}\label{DSGD_convergence}
\limsup\limits_{k\ra\infty}\frac{1}{n}\sum_{i=1}^{n}\mathbb{E}\left[\left\|\bds\theta_k^i-\bds\theta^*\right\|_2^2\right]
=  \mc{O}\left(\frac{\alpha\sigma^2}{n\mu}
+ \frac{L^2}{\mu^2}\frac{\alpha^2\sigma^2}{1-\lambda}
+ \frac{L^2}{\mu^2}\frac{\alpha^2b}{\left(1-\lambda\right)^2}\right),
\end{equation}
where~$b \triangleq \frac{1}{n}\sum_{i=1}^{n}\left\|\nabla f_i\left(\bds\theta^*\right)\right\|^2$. With a diminishing step-size~$\alpha_k = \mc{O}(\frac{1}{k})$, DSGD achieves an exact convergence~\cite{DSGD_Anit}, such that
\begin{align}\label{DSGD_diminishing}
\frac{1}{n}\sum_{i=1}^{n}\mathbb{E}\left[\left\|\bds\theta_k^i-\bds\theta^*\right\|_2^2\right] = \mc{O}\left(\frac{1}{k}\right),\qquad\forall k\geq 0.
\end{align}

\begin{rmk}\label{R2}
Comparing~\eqref{sgd_conv} to~\eqref{DSGD_convergence}, when a constant step-size~$\alpha$ is used, the mean-squared error in both SGD and DSGD decays geometrically to certain a neighborhood of~$\bth^*$, the size of which is controlled by~$\alpha$. Unlike SGD, however, the steady-state error of DSGD has an additional bias, independent of the variance~$\sigma^2$ of stochastic gradient, that comes from~$b=\frac{1}{n}\sum_{i=1}^{n}\left\|\nabla f_i\left(\bds\theta^*\right)\right\|^2$. The constant~$b$ is not zero in general and characterizes the difference between the minimizer of each local objective~$f_i$ and the global objective~$F$. This bias~$\mc{O}\big(\frac{L^2}{n\mu^2}\frac{\alpha^2b}{\left(1-\lambda\right)^2}\big)$ can be significantly large when the data distributions across all nodes are substantially heterogeneous, a scenario that commonly arises in certain IoT applications. Next, we describe a gradient tracking technique that eliminates the bias due to the term~$b$ in DSGD and thus can be considered as a more appropriate decentralized version of the centralized SGD.
\end{rmk}

\subsection{Decentralized First-Order Methods with Gradient Tracking}
\label{GT-DSGD}
To present the intuition behind the gradient tracking technique, we first recall the iterations of the Decentralized Gradient Descent (DGD) with a constant step-size in~\eqref{DGD}. Let us first assume, for the sake of argument, that at some iteration~$k$, all nodes agree on the minimizer of~$F$, i.e.,~$\bds\theta_k^i = \bds\theta^*,\forall i$. Then at the next iteration~$k+1$, we have
\begin{equation}\label{DGD_limit}
	\bds\theta_{k+1}^i = \sum_{r\in\mc N_i}w_{ir}\bds\theta^*-\alpha\nabla f_i(\bds\theta^*)=\bds\theta^*-\alpha\nabla f_i(\bds\theta^*),
\end{equation}
where~$\bds\theta^*-\nabla f_i(\bds\theta^*) \neq\bds\theta^*$, in general. In other words, the optimal~$\bds\theta^*$ is not necessarily a fixed point of~\eqref{DGD}. Of course, using the gradient~$\nabla F\left(\bds\theta_k^i\right)$ of the \textit{global} objective, instead of~$\nabla f_i\left(\bds\theta_k^i\right)$, overcomes this issue but the global gradient is not available at any node. The natural yet innovative idea of gradient tracking is to design a local iterative gradient tracker~$\mb{d}_k^i$ that asymptotically approaches the global gradient~$\nabla F\left(\bds\theta_k^i\right)$ as~$\bds\theta_k^i$ approaches~$\bth^*$
\cite{NEXT_scutari,GT_CDC,GT_Qu,DIGing,AB}. Gradient tracking is realized with the help of dynamic average consensus (DAC)~\cite{DAC}, briefly described next.
	
\newpage
In contrast to the classical average-consensus~\cite{consensus_Xiao}, which learns the average of fixed initial~states, DAC~\cite{DAC} tracks the average of time-varying signals. Formally, each node~$i$ measures a time-varying signal~$\mb{r}_k^i$ and all nodes cooperate to track the average~$\ol{\mb{r}}_k\triangleq\frac{1}{n}\sum_{i=1}^{n}\mb{r}_k^i$ of these signals. The DAC protocol is given as follows. Each node~$i$ iteratively updates its estimate~$\mb{d}_k^i$ of~$\ol{\mb{r}}_k$ as
\begin{equation}\label{DAC}
\mb{d}_{k+1}^i = \sum_{r\in\mc{N}_i}w_{ir}\mb{d}_{k}^r + \mb{r}_{k+1}^i - \mb{r}_{k}^i, \qquad k\geq0,
\end{equation}
where~$\mb{d}_0^i = \mb{r}_{0}^i, \forall i$. It is shown in~\cite{DAC} that if~$\left\|\mb{r}_{k+1}^i - \mb{r}_{k}^i\right\|_2\ra0$, we have that~$\left\|\mb{d}_{k}^i-\ol{\mb{r}}_k\right\|_2\ra0$. Clearly, in the aforementioned design of gradient tracking, the time-varying signal that we intend to track is the average of the local gradients~$\frac{1}{n}\sum_{i=1}^{n}\nabla f_i\left(\bds\theta_k^i\right)$. We thus combine DGD~\eqref{DGD} and DAC~\eqref{DAC} to obtain \textit{GT-DGD (DGD with Gradient Tracking)}~\cite{NEXT_scutari,GT_CDC,GT_Qu,DIGing,AB}, as follows:
	\begin{subequations}\label{DGT}
		\begin{align}
		\bds\theta_{k+1}^i &= \sum_{r\in\mc{N}_i}w_{ir}\bds\theta_{k}^r - \alpha\cdot\mb{d}_{k}^i, \label{DGT1}\\
		\mb{d}_{k+1}^i &= \sum_{r\in\mc{N}_i}w_{ir}\mb{d}_{k}^r + \nabla f_i\left(\bds\theta_{k+1}^i\right) - \nabla f_i\left(\bds\theta_k^i\right),\label{DGT2}
		\end{align}
	\end{subequations}
where~$\mb{d}_0^i = \nabla f_i\left(\bds\theta_0^i\right),\forall i$. Intuitively, as~$\bds\theta_k^i\ra\ol{\bds\theta}_k$ and~$\mb{d}_k^i\ra\frac{1}{n}\sum_{i=1}^n\nabla f_i\big(\bth_k^i\big)\ra\nabla F\big(\ol{\bth}_k\big)$,~\eqref{DGT1} asymptotically becomes the centralized batch gradient descent. It has been shown in~\cite{GT_Qu,DIGing,AB,DGT_NIPS} that GT-DGD converges geometrically to the global minimizer~$\bth^*$ of~$F$ under a~\textit{constant step-size} when each~$f_{i,j}\in\mc{S}_{\mu,L}$, eliminating the steady-state error~\eqref{DSGD_convergence} of DGD. 

\begin{algorithm}[!h]
	\caption{GT-DSGD: At each node~$i$}
	\begin{algorithmic}[1]
		\Require$\bds\theta_0^i$,~$\{\alpha_k\}_{k\geq0}$,~$\{w_{ir}\}_{r\in\mc N_i}$ ,~$\mb{d}_0^i=\nabla f_{s_0^i}(\bth_0^i)$, where~$s_0^i$ is chosen randomly in~$\{1,\cdots,m_i\}$
		\For{$k= 0,1,2,\cdots$}
	    \State\textbf{Update}~$\bds\theta_{k+1}^i = \sum_{r\in\mc{N}_i}w_{ir}\bds\theta_k^r- \alpha_k \mb{d}_k^i$
	    \State \textbf{Choose}~$s_{k+1}^i$ randomly in~$\{1,\cdots,m_i\}$ 
		\State \textbf{Compute} the local stochastic gradient~$\nabla f_{i,s_{k+1}^i}(\bth_{k+1}^i)$
		\State \textbf{Update}:~$\mb{d}_{k+1}^i = \sum_{r\in\mc{N}_i}w_{ir}\mb{d}_{k}^r + \nabla f_{i,s_{k+1}^i}(\bth_{k+1}^i) - \nabla f_{i,s_{k}^i}(\bth_{k}^i)$
		\EndFor
	\end{algorithmic}
\end{algorithm}
The stochastic variant of GT-DGD is derived in~\cite{DSGT}, termed as~\textit{GT-DSGD (DSGD with Gradient Tracking)}, and is formally described in Algorithm~2. Under the same assumptions of smoothness, strong-convexity, and bounded variance as in DSGD, the convergence of GT-DSGD is summarized in the following~\cite{DSGT}: with a constant step-size,~$\alpha_k=\alpha\in\left(0,\mc{O}\left(\frac{(1-\lambda)\mu}{L^2}\right)\right],\forall k$,~$\mathbb{E}[\|\bds\theta_k^i-\bds\theta^*\|_2^2]$ decays geometrically at the rate of~$\left(1-\mc{O}(\mu\alpha)\right)^k$ to a neighborhood of~$\bth^*$ such that
\begin{equation}\label{DSGT_convergence}
	\limsup\limits_{k\ra\infty}\frac{1}{n}\sum_{i=1}^{n}\mathbb{E}\left[\left\|\bds\theta_k^i-\bds\theta^*\right\|_2^2\right]
	= \mc{O}\left(\frac{\alpha\sigma^2}{n\mu}
	+ \frac{L^2}{\mu^2}\frac{\alpha^2\sigma^2}{\left(1-\lambda\right)^3}\right);
\end{equation}
with a diminishing step-size~$\alpha_k = \mc{O}(\frac{1}{k})$, we have
\begin{align}\label{DSGT_diminishing}
\frac{1}{n}\sum_{i=1}^{n}\mathbb{E}\left[\left\|\bds\theta_k^i-\bds\theta^*\right\|_2^2\right] = \mc{O}\left(\frac{1}{k}\right),\qquad\forall k\geq0.
\end{align}

\begin{rmk}
To practically implement GT-DSGD, each node needs to store its local (stochastic) gradient~$\nabla f_{i,s_k^i}(\bds\theta_k^i)$ at each time~$k$ to be used in the next iteration. Furthermore, GT-DSGD requires two consecutive rounds of communication with neighboring nodes to update the estimate~$\bth_k^i$ and the gradient tracker~$\mb{d}_k^i$, respectively. This may increase the communication burden of the network when~$\bth_k^i$ is of high dimension. \end{rmk}

\begin{rmk}\label{R4}
By comparing the convergence of DSGD in~\eqref{DSGD_convergence} and GT-DSGD in~\eqref{DSGT_convergence}, we note that under a constant step-size, GT-DSGD removes the bias~$\mc{O}\left(\frac{L^2}{\mu^2}\frac{\alpha^2b}{\left(1-\lambda\right)^2}\right)$ caused by~$b \triangleq \frac{1}{n}\sum_{i=1}^{n}\left\|\nabla f_i\left(\bds\theta^*\right)\right\|^2$ in DSGD. However, the network dependence in GT-DSGD,~$\mc{O}\left(\frac{1}{(1-\lambda)^3}\right)$, is worse than DSGD where it is~$\mc{O}\left(\frac{1}{(1-\lambda)^2}\right)$. A tradeoff here is imminent where the two approaches have their own merits depending on the relative sizes of~$b$ and~$\lambda$. Clearly, when the bias~$b$ dominates, e.g., when the data across nodes is significantly diverse, GT-DSGD achieves a lower steady-state error than DSGD. 
Under diminishing
step-sizes, DSGD and GT-DSGD have comparable performance.
Of relevance here are EXTRA~\cite{EXTRA} and Exact Diffusion~\cite{Exact_Diffusion}, both of which eliminate the bias caused by~$b$ and are built on a different principle from gradient tracking.  
\end{rmk}

\begin{rmk}\label{R5}
We note that the performance of GT-DSGD has similarities to that of the centralized SGD as in the steady-state error in both methods are completely controlled by the step-size~$\alpha$ and the variance~$\sigma^2$ of the stochastic gradient, see also Remark~\ref{R2}. Since GT-DSGD removes the bias in DSGD that comes due to the difference of the local and global objective functions, it may be considered as a more appropriate decentralized version of SGD. This argument naturally leads to the idea that one can further incorporate the centralized variance reduction techniques in the GT-DSGD framework to further improve the performance and achieve faster convergence. As we will show, adding variance reduction to GT-DSGD in fact also improves its network dependence to~$\mc{O}\left(\frac{1}{(1-\lambda)^2}\right)$. We explore these ideas next.
\end{rmk}
	
\section{Decentralized Variance-Reduced Methods with Gradient Tracking}\label{S3}
The construction of decentralized VR methods now follows from Remark~\ref{R5}. First, recall from Section~\ref{VRSGD} that the VR methods estimate the batch
gradient from randomly drawn samples. In the decentralized case, each node~$i$ thus implements VR locally to estimate its local batch gradient~$\nabla f_i$. Gradient tracking, on the other hand, estimates $\frac{1}{n}\sum_i\nabla f_i$ over the nodes and can be thought of as spatial fusion over sparse communication graphs. We now incorporate the two VR methods, SAGA and SVRG described in Section~\ref{sERM}, in the GT-DSGD framework to obtain their decentralized counterparts, called GT-SAGA~\cite{GT-SAGA} and GT-SVRG~\cite{GT-SVRG}. These algorithms are described next. Clearly, other VR approaches such as SAG and SARAH are also applicable here.

\subsection{GT-SAGA}
To implement the SAGA estimators locally, each node~$i$ maintains a gradient table that stores all local component gradients~$\{\nabla f_i(\wh{\bth}_{i,j})\}_{j=1}^{m_i}$, where~$\wh{\bth}_{i,j}$ represents the most recent iterate where the gradient of~$f_{i,j}$ was evaluated. At iteration~$k\geq0$, each node~$i$ chooses an index~$s_k^i$ randomly from~$\{1,\cdots,m_i\}$ and computes the local SAGA gradient~$\mb{g}_k^i$ as
\begin{align}
\mb{g}_{k}^{i} = \nabla f_{i,s_k^i}\big(\bth_{k}^{i}\big) - \nabla f_{i,s_k^i}\big(\wh{\bth}_{i,s_k^i}\big) + \frac{1}{m_i}\sum_{j=1}^{m_i}\nabla f_{i,j}\big(\wh{\bth}_{i,j}\big),
\end{align}
where it is straightforward to show that~$\mb{g}_k^i$ is an unbiased estimator of the local batch gradient~$\nabla f_i(\bth_k^i)$. Next, the element~$\nabla f_{i,s_k^i}(\wh{\bth}_{i,s_k^i})$ in the gradient table is replaced by~$\nabla f_{i,s_k^i}\big(\bth_{k}^{i}\big)$, while the other elements are unchanged. The gradient tracking iteration~$\mb d_k^i$ is now implemented on the estimators~$\mb g_k^i$'s over the neighboring nodes. The complete implementation of GT-SAGA~\cite{GT-SAGA} is summarized in Algorithm~\ref{GT-SAGA}. 

\begin{algorithm}
\caption{GT-SAGA at each node~$i$}
\label{GT-SAGA}
\begin{algorithmic}[1] \Require$\bds\theta_0^i$,~$\alpha$,~$\{w_{ir}\}_{r\in\mc N_i}$,~$\mb{d}_0^i=\nabla f_i(\bth_0^i)$, Gradient table $\{\nabla f_{i,j}(\wh{\bth}_{i,j})\}_{j=1}^{m_i}$ with~$\wh{\bth}_{i,j} = \bth_0^i,\forall j$.
\For{$k= 0,1,2,\cdots$}
\State\textbf{Update} {$\bth_{k+1}^i = \sum_{r\in\mc N_i}w_{ir}\bth_{k}^r - \alpha\mb{d}_{k}^{i}$;}
\State\textbf{Choose}~{$s_{k+1}^i$ randomly from~$\{1,\cdots,m_i\}$;}
\State\textbf{Compute}~{$\mb{g}_{k+1}^i = \nabla f_{i,s_{k+1}^i}\big(\bth_{k+1}^{i}\big) - \nabla f_{i,s_{k+1}^i}\big(\wh{\bth}_{i,s_{k+1}^i}\big) + \frac{1}{m_i}\sum_{j=1}^{m_i}\nabla f_{i,j}\big(\wh{\bth}_{i,j}\big)$;} \label{saga}
\State{\textbf{Replace}~$\nabla f_{i,s_{k+1}^i}\big(\wh{\bth}_{i,s_{k+1}^i}\big)$ by $\nabla f_{i,s_{k+1}^i}\big(\bth_{k+1}^{i}\big)$ in the gradient table.}
\State{\textbf{Update}~$\mb{d}_{k+1}^{i} = \sum_{r\in\mc N_i}w_{ir}\mb{d}_{k}^{r} + \mb{g}_{k+1}^i - \mb{g}_k^{i}$;}
\EndFor
\end{algorithmic}
\end{algorithm}Similar to centralized SAGA~\cite{SAGA}, GT-SAGA converges geometrically to~$\bth^*$ with a constant step-size. More precisely, assuming each~$f_{i,j}\in\mc{S}_{\mu,L}$ and choose~$\alpha = \min\left\{\mc{O}\left(\frac{1}{\mu M}\right),\mc{O}\left(\frac{m}{M}\frac{(1-\lambda)^2}{\kappa L}\right)\right\}$, where~$m=\min_i\{m_i\},M=\max_i\{m_i\}$, we have~\cite{GT-SAGA},
\begin{align}
\frac{1}{n}\sum_{i=1}^{n}\mathbb{E}\left[\left\|\bds\theta_k^i-\bds\theta^*\right\|_2^2\right] \leq R\left(1-\min\left\{\mc{O}\left(\frac{1}{M}\right),\mc{O}\left(\frac{m}{M}\frac{(1-\lambda)^2}{\kappa^2}\right)\right\}\right)^k,\qquad\forall k\geq 0,
\end{align}
where~$R> 0$ is some constant. In other words, GT-SAGA achieves~$\epsilon$-accuracy of~$\bth^*$ in 
$$\mc{O}\left(\max\left\{M,\frac{M}{m}\frac{\kappa^2}{(1-\lambda)^2}\right\}\log\frac{1}{\epsilon}\right)$$ parallel local component gradient computations.

\subsection{GT-SVRG}
GT-SVRG is a double-loop method that imitates the centralized SVRG. Each node at every outer loop computes a local batch gradient and during each inner loop performs a finite number of GT-DSGD (type) iterations, in addition to updating local gradient estimate variable~$\{\mb v_t^i\}_{i=1}^n$. As in centralized SVRG,~$\mb v_t^i$ is an unbiased estimator of the local batch gradient~$\nabla f_i(\bth_k^i)$. 
The detailed implementation of GT-SVRG is summarized in Algorithm~\ref{GT-SVRG}.
\begin{algorithm}
\caption{GT-SVRG at each node~$i$}
\label{GT-SVRG}
\begin{algorithmic}[1] 
\Require$\bds{\theta}_0^i$,~$\alpha$,~$\{w_{ir}\}_{r\in\mc N_i}$, arbitrary~$\mb{d}_0^i$ with~$\mb{v}_0^i=\mb{d}_0^i$.
\For{$k= 0,1,2,\cdots$}
\State{\textbf{Initialize}~$\ul{\bth}_{0}^i = \bds{\theta}^{i}_{k}$}
\State{\textbf{Compute}~$\nabla f_i(\ul{\bth}_0^i)=\frac{1}{m_i}\sum_{j=1}^{m_i}\nabla f_{i,j}(\ul{\bth}_0^i)$ }
\For{$t= 0,1,2,\cdots,T-1$}
\State\textbf{Update} {$\ul{\bth}_{t+1}^i = \sum_{r\in\mc{N}_i}w_{ir}\ul{\bth}_{t}^r - \alpha\cdot\mb{d}_{t}^{i}$;}
\State\textbf{Choose}~{$s_{t+1}^i$ randomly in~$\{1,\cdots,m_i\}$;}
\State\textbf{Compute}~{$\mb{v}_{t+1}^i = \nabla f_{i,s_{t+1}^i}\big(\ul{\bth}_{t+1}^i\big) - \nabla f_{i,s_{t+1}^i}\big(\ul{\bth}_0^i\big) + \nabla f_i(\ul{\bth}_0^i)$;} \label{saga}
\State{\textbf{Update}~$\mb{d}_{t+1}^i = \sum_{r\in\mc N_i}w_{ir}\mb{d}_{t}^{r} + \mb{v}_{t+1}^i - \mb{v}_{t}^i$;}
\EndFor
\State{\textbf{Set}~$\mb{d}_{0}^i = \mb{d}_{T}^i$} and~$\mb{v}_0^i = \mb{v}_T^i$
\State{Option (a):~\textbf{Set}~$\bds{\theta}_{k+1}^i = \ul\bth_{T}^i$ }
\State{Option (b):~\textbf{Set}~$\bds{\theta}_{k+1}^i = \frac{1}{T}\sum_{k=0}^{T-1}\ul\bth_{t}^i$ }
\State{Option (c):~\textbf{Set}~$\bds{\theta}_{k+1}^i$ as a random selection from~$\{\ul\bth_t^i\}_{t=0}^{T-1}$ }
\EndFor
\end{algorithmic}
\end{algorithm}

In practice, all options (a)-(c) work similarly well. For example, under option (a), it is shown in~\cite{GT-SVRG} that with~$\alpha=\mc{O}\left(\frac{(1-\lambda)^2}{\kappa L}\right)$ and~$T = \mc{O}\left(\frac{\kappa^2}{(1-\lambda)^2}\right)$, the outer loop of GT-SVRG follows:
\begin{align}
\frac{1}{n}\sum_{i=1}^{n}\mathbb{E}\left[\left\|\bds{\theta}_k^i-\bds\theta^*\right\|_2^2\right] \leq U\cdot0.9^k,   
\end{align}
where~$U>0$ is some constant. This argument implies that GT-SVRG achieves~$\epsilon$-accuracy of~$\bth^*$ in~$\mc{O}\left(\log\frac{1}{\epsilon}\right)$ outer loop iterations. Furthermore, during each inner loop, each node~$i$ computes~$m_i+2T$ local component gradients. GT-SVRG thus achieves~$\epsilon$-accuracy of~$\bth^*$ in
$$
\mc{O}\left(\left(M + \frac{\kappa^2}{(1-\lambda)^2}\right)\log\frac{1}{\epsilon}\right)
$$
parallel local component gradient computations.

\begin{rmk}
We note that both GT-SAGA and GT-SVRG have a low per-iteration computation cost and achieve geometric convergence to~$\bth^*$, i.e., they reach~$\epsilon$-accuracy of~$\bth^*$ respectively in~$\mc{O}\left(\max\left\{M,\frac{M}{m}\frac{\kappa^2}{(1-\lambda)^2}\right\}\log\frac{1}{\epsilon}\right)$ and~$\mc{O}\left(\left(M + \frac{\kappa^2}{(1-\lambda)^2}\right)\log\frac{1}{\epsilon}\right)$ parallel local component gradient computations. This makes them particularly favorable compared with DSGD and GT-DSGD when high-precision solutions are desired. Interestingly, when each node has a large data set such that~$M\approx m \gg \frac{\kappa^2}{1-\lambda^2}$, the complexities of GT-SAGA and GT-SVRG become~$\mc{O}(M\log\frac{1}{\epsilon})$, independent of the network, and are~$n$ times faster than that of centralized SAGA and SVRG. Clearly, in this ``big-data" regime, GT-SAGA and GT-SVRG act effectively as a means for parallel computation and achieve linear speed-up compared with their centralized counterparts.
\end{rmk}

\begin{rmk}It can also be observed that when data samples are distributed over the network in a highly unbalanced way, i.e.,~$\frac{M}{m}$ is very large, it may appear that GT-SVRG achieves a lower complexity than GT-SAGA. However, from a practical implementation point of view, an unbalanced data distribution may lead to a longer computation time in GT-SVRG. This is due to the number of local gradient computations required at the end of each inner loop especially for nodes with large number of data samples. GT-SVRG consequently cannot execute the next inner loop before all nodes finish the local batch gradient computation, leading to an overall increase in runtime. Clearly, there is an inherent trade-off between network synchrony, latency, and the storage of gradients as far as the relative implementation complexities of GT-SAGA and GT-SVRG are concerned. If each each node is capable of storing all local component gradients, then GT-SAGA may be preferable due to its flexibility of implementation. On the other hand, for large-scale optimization problems where each node holds a very large number of data samples, storing all component gradients may be infeasible and GT-SVRG may be preferable.
\end{rmk}

\begin{rmk}
Existing decentralized VR methods include DSA~\cite{DSA} that combines EXTRA~\cite{EXTRA} with SAGA~\cite{SAGA}, diffusion-AVRG that combines exact diffusion~\cite{Exact_Diffusion} and AVRG~\cite{AVRG}, DSBA~\cite{DSBA} that adds proximal mapping~\cite{point-SAGA} to each iteration of DSA, and ADFS~\cite{ADFS} that applies an accelerated randomized proximal coordinate gradient method~\cite{APCG} to the dual formulation of Problem P3. We note that in large-scale scenarios where~$M\approx m$ is very large, both~GT-SAGA and~GT-SVRG improve upon the convergence rate of these methods in terms of the joint dependence on~$\kappa$ and~$M\approx m$, with the exception of DSBA and ADFS. Both DSBA and ADFS achieve better iteration complexity, however, at the expense of computing the proximal mapping of a component function at each iteration.	Although the computation of this proximal mapping is efficient for certain function classes, it can be very expensive for general functions.
\end{rmk}

\newpage
\section{Numerical Illustrations}\label{S5}
\begin{figure*}[!h]
		\centering
		\subfigure{\includegraphics[width=2.9in]{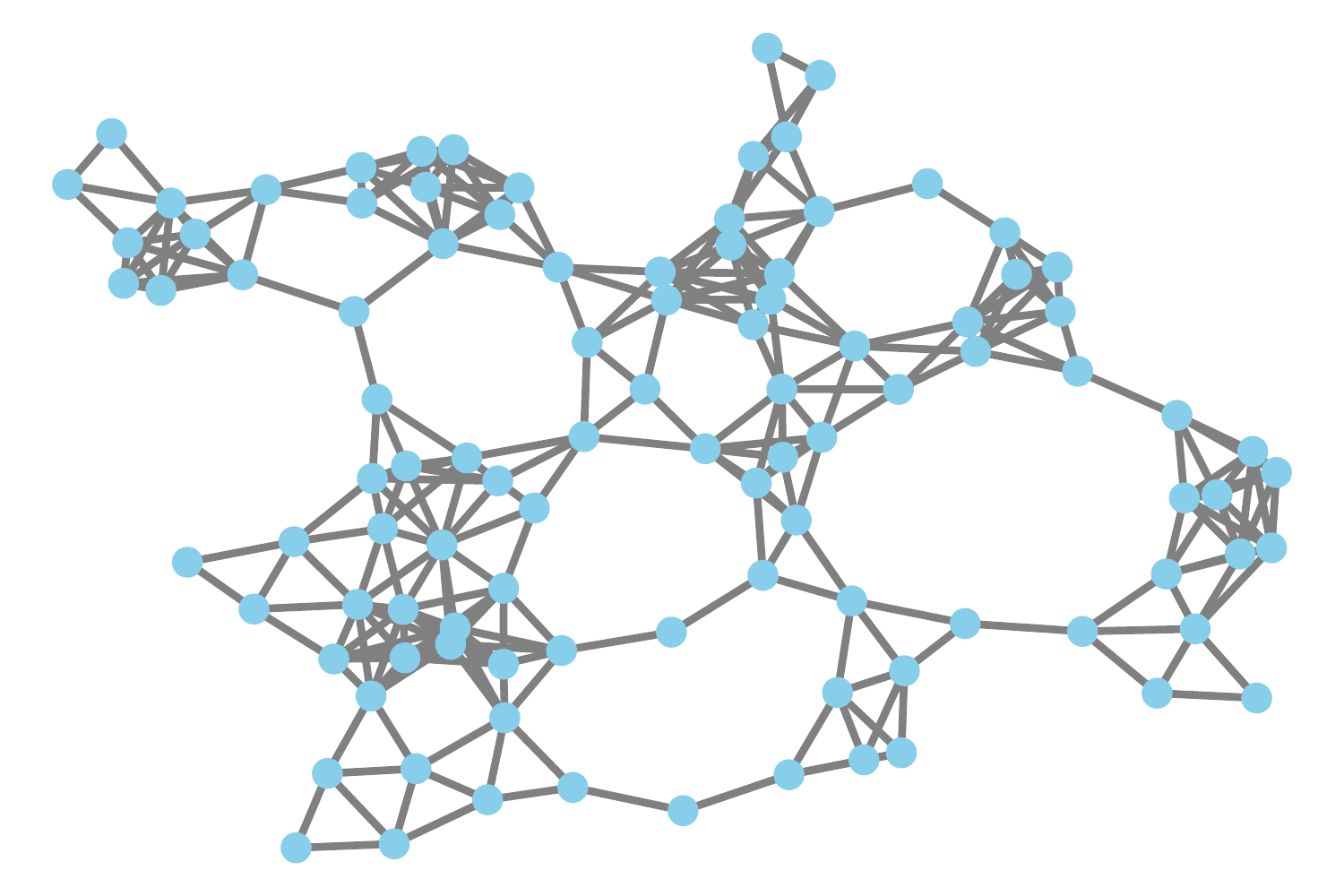}}
		\caption{A sparsely-connected geometric graph over which the nodes communicate with each other.}
		\label{network}
\end{figure*}

In this section, we present numerical experiments to illustrate the convergence properties of the consensus-based stochastic optimization algorithms presented in this article, i.e., DSGD, GT-DSGD, GT-SAGA and GT-SVRG. We compare these methods with the help of decentralized training of a regularized logistic regression model~\cite{PRML}, that is smooth and strongly-convex, to classify the hand-written digits~$\{3,8\}$ from the MNIST dataset. The digit images~$\{3,8\}$ are represented by feature vectors in~$\mathbb{R}^{784}$ that have been normalized to have zero-mean and a standard deviation of~$1$. We randomly generate a connected and undirected graph of~$100$ nodes using the nearest-neighbor rule, i.e., two nodes are connected only if they are in a certain close proximity, a particular visualization of this graph is shown in Fig.~\ref{network}. This type of connectivity commonly arises in large-scale IoT applications where devices have limited communication capabilities. The doubly-stochastic weight matrix associated with the network, which is required for the implementation of the algorithms, is generated using the Metropolis method~\cite{tutorial_nedich}. In our setup, each node~$i$ holds~$m_i=10$ training data samples,~$\left\{\mb{x}_{i,j},y_{i,j}\right\}_{j=1}^{m_i}\subseteq\mathbb{R}^{784}\times\left\{-1,+1\right\}$, where~$\mb{x}_{i,j}$ is the feature vector and~$y_{i,j}$ is the corresponding binary label. We make a further restriction that the training data samples at each node belong to only one class, either~$3$ or~$8$. In other words, no node can train a proper classifier by only using its own local batch data; clearly, the local~$f_i$'s are significantly different from the global~$F$. To train a valid classifier, the networked nodes must cooperate to solve the following logistic regression problem:
	\begin{equation}
	\operatorname*{min}_{\mb{b}\in\mathbb{R}^{784},\:c\in\mathbb{R}}F(\mb{b},c) = \frac{1}{n}\sum_{i=1}^{n}\frac{1}{m_i}\sum_{j=1}^{m_i}\ln\left[1+\exp\left\{-(\mb{b}^\top\mb{x}_{ij}+c)y_{ij}\right\}\right]+\frac{\lambda}{2}\|\mb{b}\|_2^2. \nonumber
	\end{equation}
Clearly,~$\bth=[\mb b^\top c]^\top$. For comparison, we plot the average residual~$\frac{1}{n}\sum_{i=1}^{n}\|\bds\theta_k^i-\bds\theta^*\|_2^2$ across all nodes versus the number of local epochs (number of effective passes of local data batch). The hyper-parameters for the algorithms in question are manually optimized.    

\begin{figure*}[!h]
		\centering
		\subfigure{\includegraphics[width=2.85in]{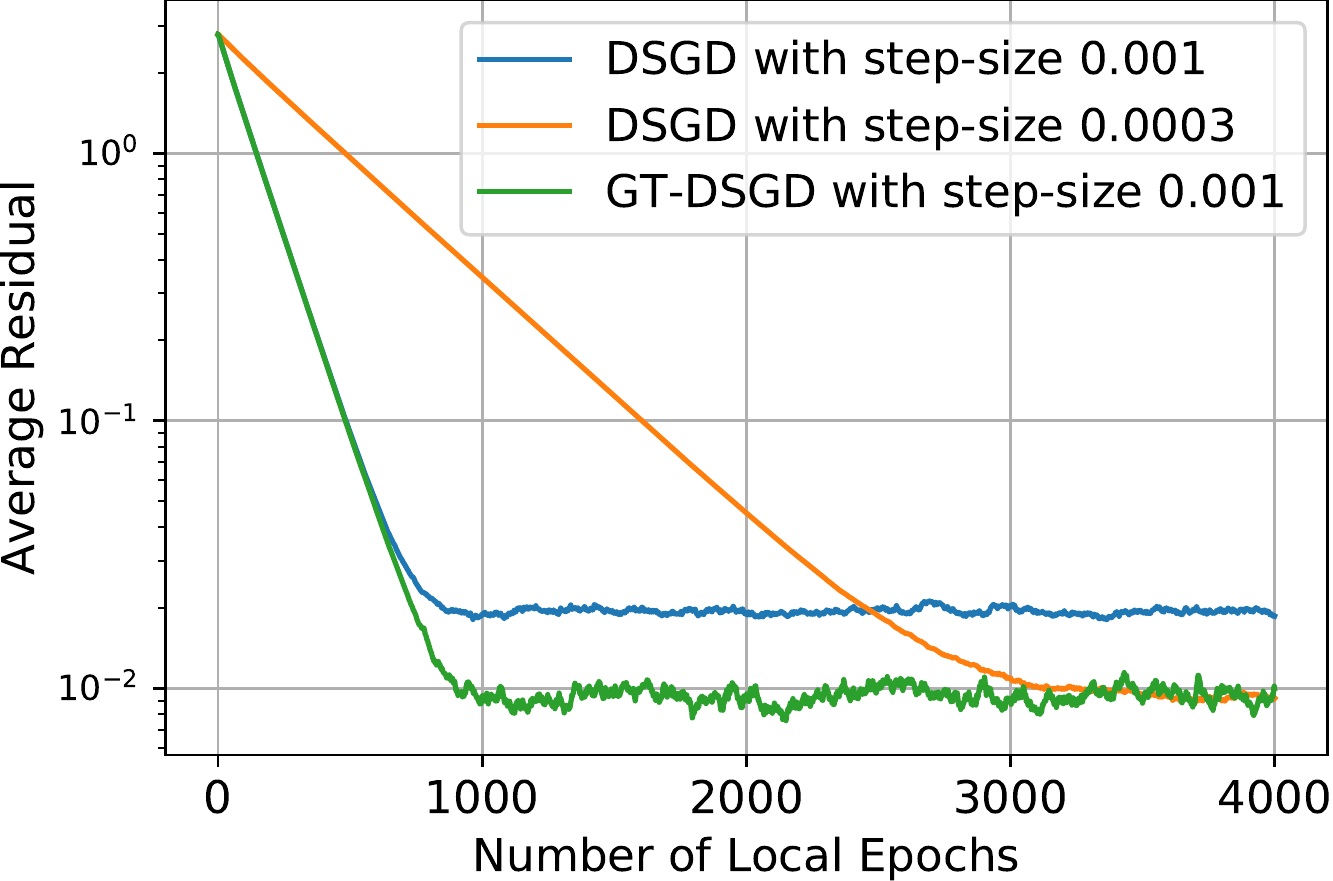}}
		\caption{Comparison of DSGD and GT-DSGD under different constant step-sizes.}
		\label{DSGD_GT}
\end{figure*}

We first compare the performance of DSGD and GT-DSGD with constant step-sizes and observe that the numerical results shown in Fig.~\ref{DSGD_GT} are consistent with Remark~\ref{R4}, i.e., under a constant step-size, DSGD has a larger steady-state error compared with GT-DSGD when the minimizers of the local and the global objective functions are significantly different (recall that each node has image data for only one digit). To achieve a smaller steady-state error as that of GT-DSGD, one needs to apply a smaller step-size in DSGD that may lead to a slower convergence~rate. 

\begin{figure*}[!h]
		\centering
		\subfigure{\includegraphics[width=2.85in]{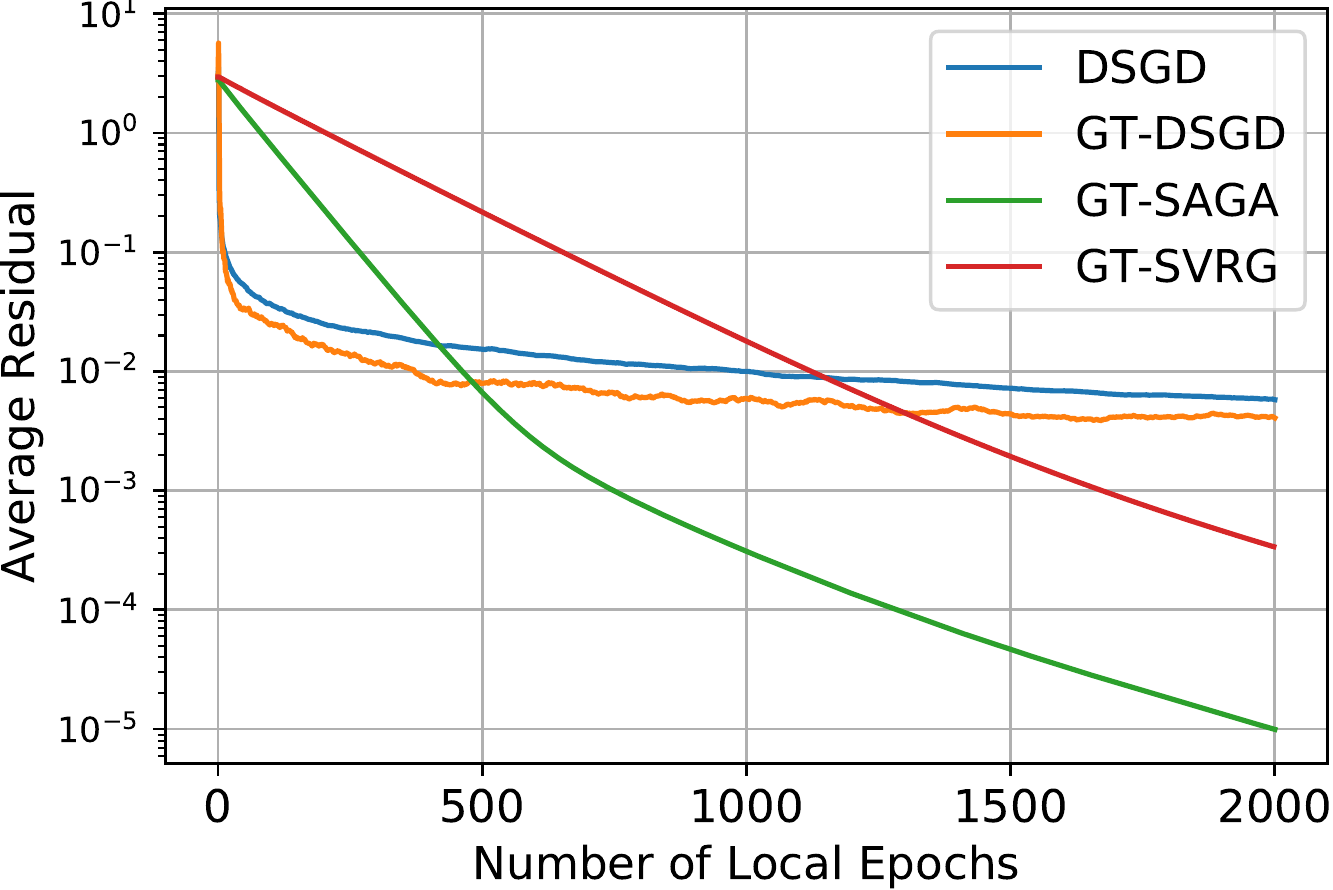}}\hspace{1cm}
		\caption{Comparison between all algorithms: DSGD with~$\alpha_k = \frac{2}{k+1}$, GT-DSGD with~$\alpha_k = \frac{3}{k+1}$, GT-SAGA with~$\alpha = 6\times 10^{-5}$ and GT-SVRG with inner-loop length~$T=10$ and~$\alpha = 5\times 10^{-5}$.}
		\label{all}
\end{figure*}

Next, we compare the performance of DSGD, GT-DSGD, both with diminishing step-sizes to ensure exact convergence, GT-SAGA, and GT-SVRG  in~Fig.~\ref{all}. It can be observed that all four algorithms are effective for the training problem in question and may be favorable in different regimes. DSGD and GT-DSGD, for example, make very fast progress in the first few epochs and then slow down significantly, which makes them more suitable for problems where low-precision solutions suffice. On the contrary, as the iterations proceed, GT-SAGA and GT-SVRG exhibit fast geometric convergence and provide highly-accurately solutions in much less iterations than DSGD and GT-DSGD. We note that GT-SAGA achieves faster convergence rate than GT-SVRG, however, with the requirement of storing the latest copy of all component gradients.  
	
\newpage
\section{Extensions and Discussion}\label{S4}
We now discuss some recent progress on several key aspects of decentralized optimization relevant to the first-order stochastic approaches described in this article. 

\textbf{Directed Graphs: }The methods described in this article are restricted to undirected graphs. In practice, however, bidirectional communication may not always be preferable or even achievable, e.g., when the nodes have non-uniform communication ranges, or, when certain communication links are severed to save on communication costs. Such scenarios lead to~\textit{directed graphs} where the main challenge is that the underlying weight matrix~$W$ can be either row-stochastic (RS) or column-stochastic (CS), but cannot be doubly-stochastic (DS), in general. The doubly-stochasticity of the weight matrix is essential for the convergence of the algorithms presented in this article. In particular, consensus cannot be reached with CS weights; and with RS weights, the nodes agree albeit on a sub-optimal solution; see~\cite{FROST} for a detailed discussion. A well-studied solution to this issue is based on the push-sum (type) algorithms~\cite{push_sum} that enable consensus with non-DS weights with the help of eigenvector estimation. Combining push-sum respectively with 
DSGD~\cite{DSGD_nedich,diffusion_Chen}, EXTRA~\cite{EXTRA}, and~GT-DGD~\cite{GT_CDC,GT_Qu,DIGing} leads to SGP~\cite{SGP_ICML}, DEXTRA~\cite{DEXTRA}, and ADD-OPT~\cite{ADD-OPT} that are applicable to arbitrary directed graphs. A similar idea is used in FROST~\cite{FROST} to implement decentralized optimization with RS weights. 

The issue with push-sum based extensions is that they require eigenvector estimation, which in itself could deteriorate the performance of the underlying algorithms. More recently, it is shown that GT-DGD~\eqref{DGT} is a special case of the AB algorithm~\cite{AB,push-pull} that employs RS weights in~\eqref{DGT1} while CS weights in~\eqref{DGT2}, and thus is immediately applicable to arbitrary directed graphs. 
The AB framework naturally leads to stochastic optimization with gradient tracking over directed graphs, see SAB~\cite{SAB} that extends GT-DSGD to directed graphs, and further opens the possibility to extend GT-SAGA and GT-SVRG to their directed counterparts. 

\textbf{Communication and computation aspects: }Communication complexity is an important aspect of decentralized optimization since communication can potentially become a bottleneck of the system when nodes are frequently transmitting high-dimensional vectors (model parameters) in the network. Different communication aspects~\cite{DOPT_lan}, communication/computation tradeoffs~\cite{tutorial_nedich}, and various quantization techniques~\cite{QDGD,QDGD2} have been studied with existing decentralized methods in an attempt to efficiently manage the resources at each node. 

\textbf{Master-worker architectures:} The problems described in this article have experienced a significant research activity recently because of their direct applicability to many large-scale training problems in machine learning~\cite{DGD_NIPS,SGP_ICML}. Since these applications are typically hosted in controlled settings, e.g., data centers with highly-sophisticated communication and a large number of highly-efficient computing clusters, master-worker architectures and parameter-server~models have become popular. In such architectures, see Fig.~\ref{decentralized} (left), a central master maintains the~current model parameters and communicates strategically with the workers, which individually hold a local batch of the total training data. The basic idea is that the master pushes the current model~$\bds\theta_k$ to the workers, each of which computes a stochastic gradient at~$\bds\theta_k$ using a random subset of its own local data; the master then pulls the stochastic gradients from the workers and updates the model. Various programming models and several variants of master-worker configurations have been proposed, such as MapReduce, All-Reduce, and federated learning~\cite{Federated_learning}, that are tailored for specific computing needs and environments. We emphasize that, on the contrary, the motivation behind consensus-based decentralized methods comes from the scenarios where communication among the nodes is ad hoc and unstructured and specialized topologies are not available. 

\vspace{-0.12cm}
\section{Conclusions}\label{S6}
In this article, we discuss general formulation and solutions for decentralized consensus-based stochastic optimization. Compared with traditional master-worker architectures, consensus-based optimization eliminates the need of a central coordinator and admits sparse and flexible peer-to-peer communication that enjoys reduced communication cost at each node, is more robust in ad-hoc and adversarial environments, and is further applicable to the emergent IoT applications where the nodes have resource-constraints and limited communication capabilities. We discuss several fundamental algorithmic frameworks with a focus on gradient tracking and variance-reduction methods. For all of the related algorithms, we provide a detailed discussion on their convergence rates, properties, and comparisons with a particular focus on smooth and strongly-convex objective functions. An important line of future work in the field of consensus-based optimization for machine learning is to analyze existing methods and develop new techniques for general non-convex objectives, given the tremendous success of~deep neural networks. 	

\vspace{-0.2cm}
\bibliographystyle{IEEEbib}
\bibliography{SPM.bib}
\end{document}